\documentclass[sigconf]{acmart}

\AtBeginDocument{%
  \providecommand\BibTeX{{%
    \normalfont B\kern-0.5em{\scshape i\kern-0.25em b}\kern-0.8em\TeX}}}

\copyrightyear{2021}
\acmYear{2021}
\setcopyright{acmcopyright}
\acmConference[MM '21] {Proceedings of the 29th ACM International Conference on Multimedia}{October 20--24, 2021}{Virtual Event, China.}
\acmBooktitle{Proceedings of the 29th ACM International Conference on Multimedia (MM '21), Oct. 20--24, 2021, Virtual Event, China}
\acmPrice{15.00}
\acmDOI{10.1145/3474085.3475490}
\acmISBN{978-1-4503-8651-7/21/10}

\usepackage[normalem]{ulem}
\usepackage{pifont}
\usepackage{enumitem}
\usepackage{multirow}
\usepackage{wrapfig}
\newcommand{\cmark}{\ding{51}}%
\newcommand{\xmark}{\ding{55}}%
\newcommand*\chancery{\fontfamily{pzc}\selectfont}

\begin{document}
\fancyhead{}
\title{WAS-VTON: Warping Architecture Search for Virtual Try-on Network}

\author{Zhenyu Xie}
\author{Xujie Zhang}
\author{Fuwei Zhao}
\affiliation{%
  \institution{Shenzhen Campus of Sun Yat-Sen University}
  \city{Shenzhen}
  \state{Guangdong}
  \country{China}
}



\author{Haoye Dong}
\affiliation{%
  \institution{Sun Yat-Sen University}
  \city{Guangzhou}
  \state{Guangdong}
  \country{China}
}

\author{Michael C. Kampffmeyer}
\affiliation{%
  \institution{UiT The Arctic University of Norway}
  \state{Tromsø}
  \country{Norway}
}

\author{Haonan Yan}
\affiliation{%
  \institution{Momo Technology Company Limited}
  \state{Beijing}
  \country{China}
}

\author{Xiaodan Liang}
\authornote{Xiaodan Liang is the corresponding author. Our code is available at \url{https://github.com/xiezhy6/WAS-VTON}.}
\affiliation{%
  \institution{Shenzhen Campus of Sun Yat-Sen University}
  \institution{DarkMatter AI Research}
  \city{Shenzhen}
  \state{Guangdong}
  \country{China}
}
\email{xdliang328@gmail.com}




\renewcommand{\shortauthors}{Zhenyu Xie, et al.}
\begin{abstract}
  Despite recent progress on image-based virtual try-on, current methods are constraint by shared warping networks and thus fail to synthesize natural try-on results when faced with clothing categories that require different warping operations. In this paper, we address this problem by finding clothing category-specific warping networks for the virtual try-on task via Neural Architecture Search (NAS). We introduce a NAS-Warping Module and elaborately design a bilevel hierarchical search space to identify the optimal network-level and operation-level flow estimation architecture. Given the network-level search space, containing different numbers of warping blocks, and the operation-level search space with different convolution operations, we jointly learn a combination of repeatable warping cells and convolution operations specifically for the clothing-person alignment. Moreover, a NAS-Fusion Module is proposed to synthesize more natural final try-on results, which is realized by leveraging particular skip connections to produce better-fused features that are required for seamlessly fusing the warped clothing and the unchanged person part. We adopt an efficient and stable one-shot searching strategy to search the above two modules. Extensive experiments demonstrate that our WAS-VTON significantly outperforms the previous fixed-architecture try-on methods with more natural warping results and virtual try-on results.
\end{abstract}

\ccsdesc[500]{Computing methodologies~Computer vision tasks}
\keywords{Generative Model; Image Synthesis; Virtual Try-on; Neural Network Search}

\begin{teaserfigure}
  \vspace{-5mm}
  \includegraphics[width=\textwidth]{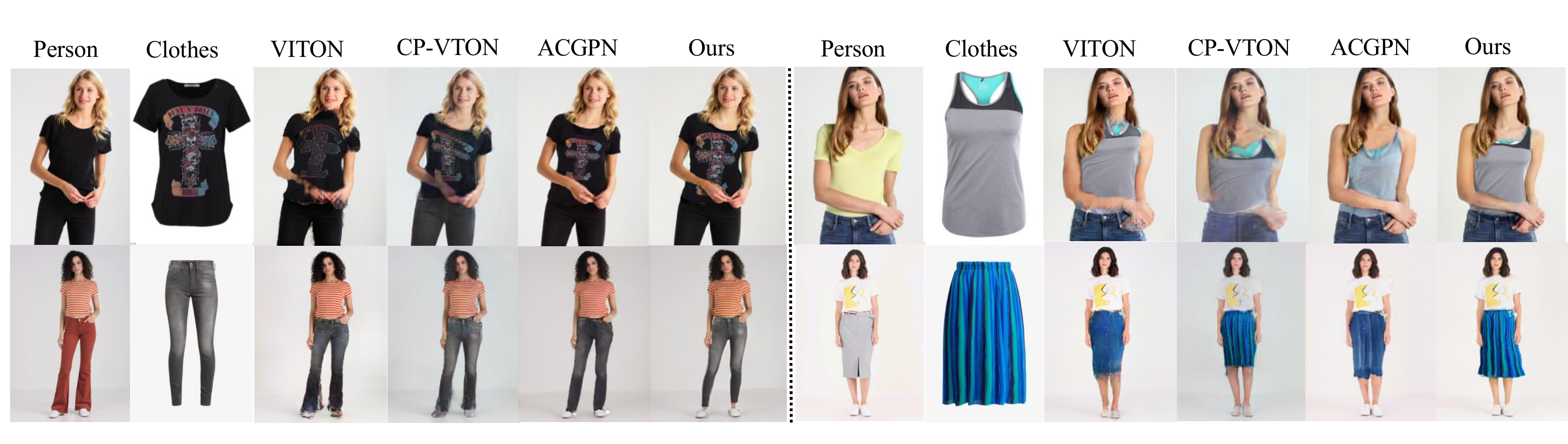}
  \vspace{-7mm}
  \caption{Comparison between our WAS-VTON and recent state-of-the-art methods. With the searched architecture specifically designed for the try-on task, WAS-VTON can perform high-quality full-body virtual try-on and achieves better results than previous architecture-fixed approaches.}
  \label{fig:teaser}
\end{teaserfigure}

\maketitle

\section{Introduction}
Image-based virtual try-on aims to transfer the desired clothing onto a particular person and has been rapidly developed due to the great progress in generative models~\cite{goodfellow2014gan,Mirza2014ConditionalGA,pix2pix2017,wang2018pix2pixHD} as well as its large practical value.
The core challenge of virtual try-on is to find a proper warping strategy that can account for spatial misalignment between source and target clothing regions.
Existing works~\cite{xintong2018viton,bochao2018cpvton,yun2019vtnfp,xintong2019clothflow,Matiur2020cpvton+,han2020acgpn,thibaut2020swuton} typically solve this challenge by mimicking nonrigid clothing deformations via TPS-based methods or flow-based methods.
TPS-based methods estimate parameters of the Thin Plate Spline (TPS)~\cite{bookstein1989TPS} transformation to warp the in-shop clothing onto the human body. While results are overall good, the TPS transformation, due to its limited degrees of freedom (e.g., $2\times5\times5$ as in~\cite{xintong2018viton,bochao2018cpvton}), tends to fail when faced with large clothing deformations or severe occlusions (see the 1st row in Fig.~\ref{fig:teaser}).  To address these issues, flow-based methods~\cite{xintong2019clothflow} enlarge the deformation space by estimating a dense flow field (e.g., $2\times256\times256$ in~\cite{xintong2019clothflow}) for higher flexibility when dealing with complex clothing deformation. 
Compared to the TPS transformation, 
flow-based approaches operate on the finest element in image space, i.e., the pixel-level transformation between source and target clothing regions. They, thereby, capture the spatial correspondence more accurately and consequently help realistically synthesizing the target image. 

\begin{figure}[t]
  \centering
  \includegraphics[width=1.0\hsize]{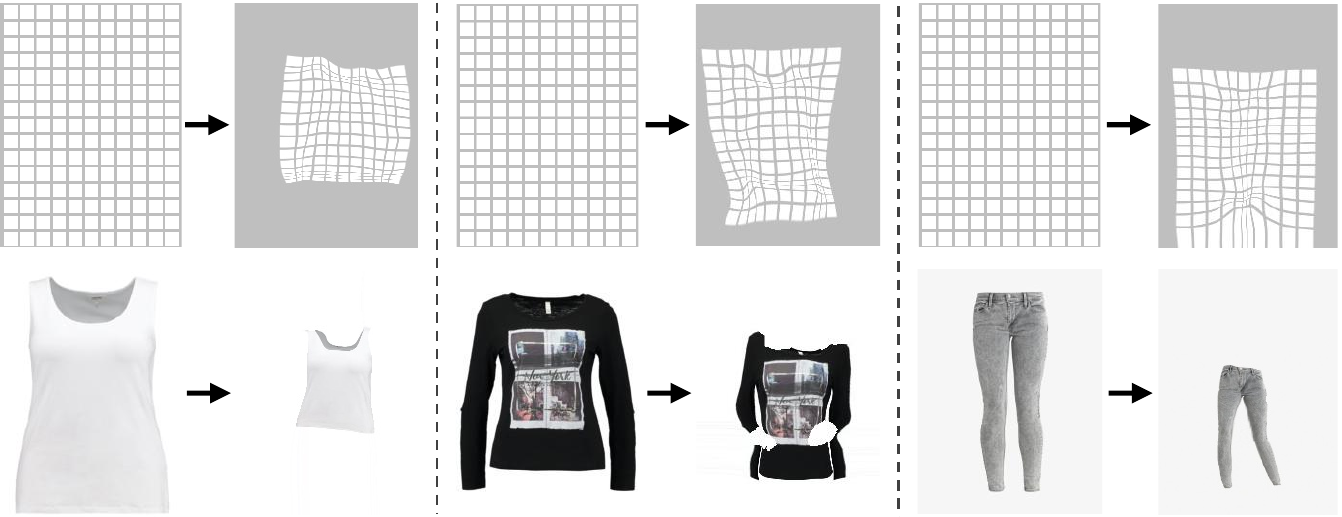}
  \vspace{-4mm}
  \caption{Deformation comparison for different clothing categories during the try-on procedure. Please zoom in for more details.}
  \vspace{-7mm}
  \label{fig:defomation_comparison}
\end{figure}

However, despite the increased deformation capacity, flow-based methods consistently use a fixed warping architecture for different clothing categories. We argue that this category-agnostic design leads to considerable inefficiency and confines possible deformations to a locally entangled subspace, resulting in inferior virtual try-on results. 
Note, the complexity of clothing deformations tends to vary with the clothing category during the try-on process. For instance, the deformation of the vest in Fig.~\ref{fig:defomation_comparison} is close to an affine transformation, while the deformations of the long sleeve and pants are much more complicated due to the commonly locally non-rigid transformations.
Motivated by this phenomenon, we want to find a flexible and efficient category-specific warping architecture to handle the deformation discrepancy among different clothing types.

Naively handcrafting such an architecture would be cumbersome and error-prone, since the flow-based warping network is normally inner cascaded~\cite{xintong2019clothflow} and it is thus non-trivial to manually guide the flow toward the correct category-specific deformation subspace. 
Inspired by the recent successes of Neural Architecture Search (NAS)~\cite{nas_survey} to identify optimal neural network architectures for computer vision tasks such as image classification~\cite{liu2018dasrts,bender2018oneshot,zichao2019singlepath,xu2019pc-darts,li2020dna,tan2019efficientnet} and semantic segmentation~\cite{liu2019auto-deeplab,zhang2019cas}, we propose an elaborately designed bilevel hierarchical search space that is specifically tailored to the virtual try-on task. We explore the search space using a highly efficient searching rule and modularize this algorithm in our \emph{NAS-Warping Module}.

To be specific, we design a network-level search space and an operation-level search space for the NAS-Warping Module. The former consists of five multi-scale warping cells, each of which is composed of one to three warping blocks. The operation-level search space contains four types of convolution operations within each warping block. The goal of this module is to jointly learn a combination of repeatable warping cells and convolution operations specifically for the clothing warping stage. 

Besides image warping, the other key step in the virtual try-on pipeline is to seamlessly fuse the warped clothing and the unchanged person part by an U-Net-like fusion network~\cite{xintong2018viton,bochao2018cpvton,han2020acgpn}. As illustrated in~\cite{pix2pix2017}, the skip connections in the U-Net play an important role in such an image-to-image translation problem. We therefore further propose a \emph{NAS-Fusion Module} that has a network-level search space containing skip connections not only between the same feature scales as in previous try-on approaches~\cite{xintong2018viton,bochao2018cpvton,yun2019vtnfp} but also between different feature scales. 
Moreover, the NAS-Fusion Module also comprises an operation-level search space with different convolution operation for the each network layer.
The purpose of the NAS-Fusion Module is to improve the image translation network in~\cite{pix2pix2017} by finding the optimal skip connections to fuse features and generate the final try-on results.

To support these two searchable modules, WAS-VTON also utilizes a preparatory \emph{Partial Parsing Module} to predict the semantic label of the replaced region, providing a target clothing mask that is used for estimating the flow in the NAS-Warping Module and yields instructive guidance for the NAS-Fusion Module to better delineate clothing and skin parts.

To demonstrate the effectiveness of our WAS-VTON, we conduct comprehensive experiments on the virtual try-on benchmark VITON~\cite{xintong2018viton}. We compare warped results and the final try-on results of WAS-VTON with three existing state-of-the-art virtual try-on methods~\cite{xintong2018viton,bochao2018cpvton,han2020acgpn} and four carefully designed baselines.
Results evidence the superiority of WAS-VTON over other architecture-fixed methods to obtain more natural warped clothes and more photo-realistic try-on results (see Fig.~\ref{fig:teaser}).

Overall, our contributions are three-fold:
\begin{itemize}[itemsep=1pt,topsep=1pt]
\item For the first time, we leverage NAS for the virtual try-on task to search clothing category-specific warping networks for different clothing categories
and an image fusion network for a more natural synthesis of the try-on results.
\item We elaborately design a novel two-level search space for both the NAS-Warping Module and the NAS-Fusion Module, namely, a network-level and operation-level search space to identify the optimal network for the flow-based virtual try-on task.
\item With the searched networks, our WAS-VTON outperforms the existing state-of-the-art try-on methods on both the clothing deformation and the 
final try-on results.
\end{itemize}

\section{Related Work}
\textbf{Virtual Try-on.}
Virtual try-on has seen a surge of interest in image generation and computer graphics research fields. 
3D graphics-based try-on methods~\cite{robert2002animation,r2003clothsimulation,selle2009hrcloth,peng2012drape,fabian2014scs,stoll2010video,gerard2017clothcap,lahner2018deepwrinkles,bhatnagar2019mgn} 
usually rely on intensive computational, manual parameter tuning, or extra facilities, making them unaffordable in practical applications.
To reduce the expensive deployment cost for virtual try-on system, recent efforts have been focusing on image-based methods~\cite{xintong2018viton,bochao2018cpvton,yun2019vtnfp,xintong2019clothflow,Matiur2020cpvton+,han2020acgpn,fancato2020viton-gt,thibaut2020swuton,neuberger2020oviton,dong2019mgvton}, which directly transfer a clothing image onto a reference person image.
Among them, VITON~\cite{xintong2018viton} and CP-VTON~\cite{bochao2018cpvton} model the clothing deformation via TPS~\cite{bookstein1989TPS} transformation and establish a try-on framework consisting of a warping module and a fusion module, building the foundation for several follow-up methods.
VITON-GT~\cite{fancato2020viton-gt} incorporates affine and TPS transformation in a two-stage warping module to exploit more accurate deformation.
VTNFP~\cite{yun2019vtnfp} makes a first attempt to utilize human parsing as guidance for synthesizing more precise body parts.
ACGPN~\cite{han2020acgpn} introduces a second-order difference constraint on the TPS parameters to stabilize the warping operation and proposes an adaptive parsing scheme to obtain more precise human parsing. 
Instead of using the TPS, ClothFlow~\cite{xintong2019clothflow} estimate the clothing flow to delineate the pixel-wise transformation between the source and the target clothing, achieving more flexible warping results.
However, both the TPS-based and flow-based methods neglect the deformation discrepancy among different clothing categories and learn a architecture-fixed warping network for various types of clothing deformations.
Besides, most methods above directly use the standard UNet for fusing the warped clothing and the person image, which is not the optimal choice for the virtual try-on task, as demonstrated by our experiments.
Our proposed WAS-VTON, instead, utilizes the NAS algorithm to search a category-specific warping network for different clothing categories and an optimal fusion network specifically for the try-on task.

\textbf{Neural Architecture Search.}
Neural Architecture Search (NAS) has been utilized to search the optimal network architecture for numerous computer vision problems, such as image classification~\cite{liu2018dasrts,bender2018oneshot,zichao2019singlepath,xu2019pc-darts,li2020dna,tan2019efficientnet}, object detection~\cite{peng2019clod,ghiasi2019nasfpn,jing2020sanas,wang2020nasfcos,wang2020efficientdet}, image segmentation~\cite{liu2019auto-deeplab,zhang2019cas}, and image synthesis~\cite{gao2019adversarialnas,gong2019autogan}.
In general, the NAS algorithms need to solve the problems of \textit{architecture search} and \textit{weight optimization}. Early NAS works use reinforcement learning~\cite{barret2016naslr,barret2018lta} or evolution algorithms~\cite{xie2017genetic} to solve these two problems sequentially, which require prohibitively high computation cost for large search spaces or huge datasets. 

To accelerate the NAS procedure, a strategy called \textit{weight sharing} has been proposed to encode the whole search space as a supernet. Once the supernet is fully trained, the optimal architecture can be sampled from it. The sampled sub-network inherits the weights from the supernet and only requires task-specific fine-tuning.
An efficient solution for such strategy is the one-shot paradigm, in which the supernet training and architecture search are conducted independently~\cite{bender2018oneshot,zichao2019singlepath,li2020dna}. To train the sub-network fairly, these kinds of methods usually train the supernet by path dropout or path sampling. 
The searched results of one-shot NAS are more stable than other approaches like~\cite{liu2018dasrts,xu2019pc-darts,gao2019adversarialnas,lahner2018deepwrinkles} of which the 
architecture search and the network weights learning are deeply coupled. 

In this work, we make the first attempt to exploit NAS for the virtual try-on task and propose a dynamic Warping Architecture Search method for Virtual Try-on Networks (WAS-VTON). To balance the searching time and the stability, our WAS-VTON inherits the one-shot NAS framework. Different from previous NAS-based image synthesis works~\cite{gao2019adversarialnas,gong2019autogan}, which utilize the fixed network architecture adapted from the image matching~\cite{cnngeometric} and translation tasks~\cite{pix2pix2017}, our WAS-VTON aims to find the optimal architecture for the core modules of the virtual try-on task.

\begin{figure*}[t]
  \centering
  \includegraphics[width=1.0\hsize]{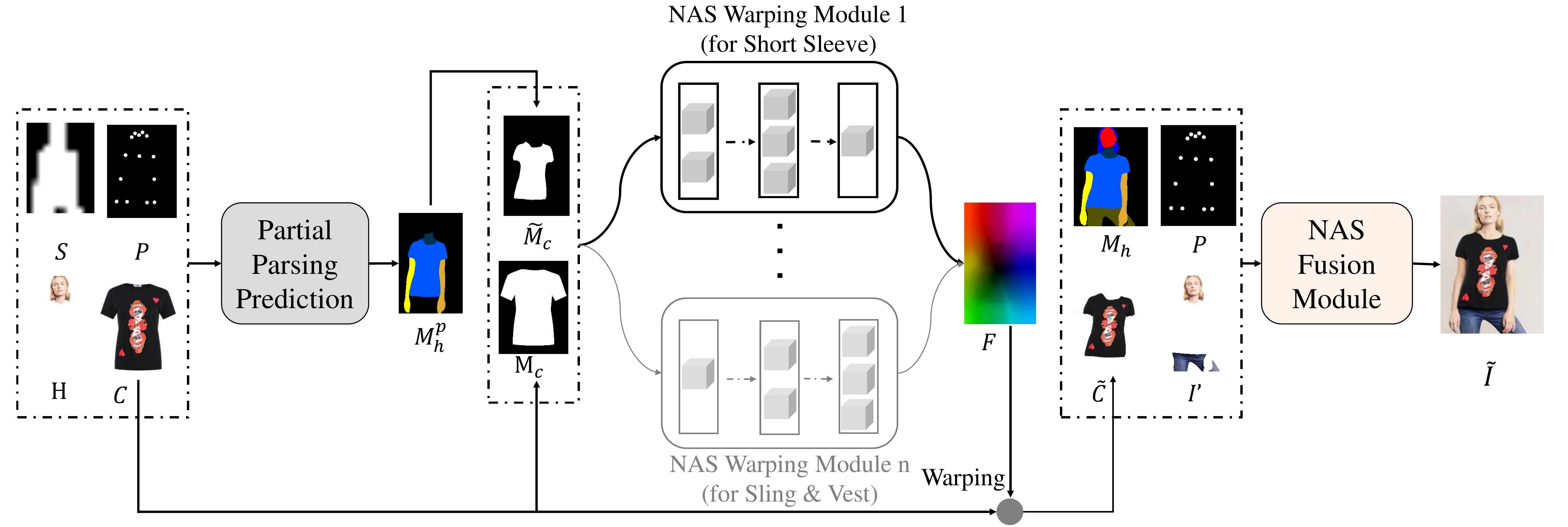}
  \vspace{-4mm}
  \caption{Framework of our WAS-VTON. The partial parsing prediction module takes the body shape $S$, the person pose $P$, the head image $H$, and the flat clothing $C$ as inputs and predicts the partial parsing $M_h^p$. Given the source clothing mask $M_c$ and target clothing mask $\tilde{M}_c$, the NAS-Warping module selects a clothing category-specific warping network to predict the clothing flow $F$, which is used to warp the flat clothing $C$ to target shape $\tilde{C}$. The NAS-Fusion module takes the human parsing $M_h$, the person pose $P$, the warped clothing $\tilde{C}$, and the partial person image $I'$ as inputs to synthesize the try-on result $\tilde{I}$.}
  \vspace{-3mm}
  \label{fig:framework}
\end{figure*}

\section{WAS-VTON}~\label{tab:method}
Synthesizing photo-realistic virtual try-on results heavily depends on the precise clothing deformation and the seamless fusion of the warped clothing and the person image.
Therefore, our proposed WAS-VTON is composed of three modules, the partial parsing prediction module, the NAS-Warping module, and the NAS-Fusion module. 
The partial parsing prediction module predicts the partial body segmentation map including neck (resp., shoes), arms (resp., legs) and the upper (resp., lower) clothing region to be changed during the upper-body (resp., lower-body) try-on process, which provides necessary information for the NAS-Warping module and guides the NAS-Fusion module to predict a precise fusion mask as well as synthesize different body parts accurately. 
The NAS-Warping module searches clothing category-specific warping networks, which use different numbers of warping operations to adapt to various degrees of deformation caused by the different clothing categories.
The NAS-Fusion module searches particular skip-connections in the encoder-decoder network to learn more accurate features for the fusion of the warped clothing and the person image. 
The overall framework of WAS-VTON is displayed in Fig.~\ref{fig:framework}.

\begin{figure}[t]
  \centering
  \includegraphics[width=1.0\hsize]{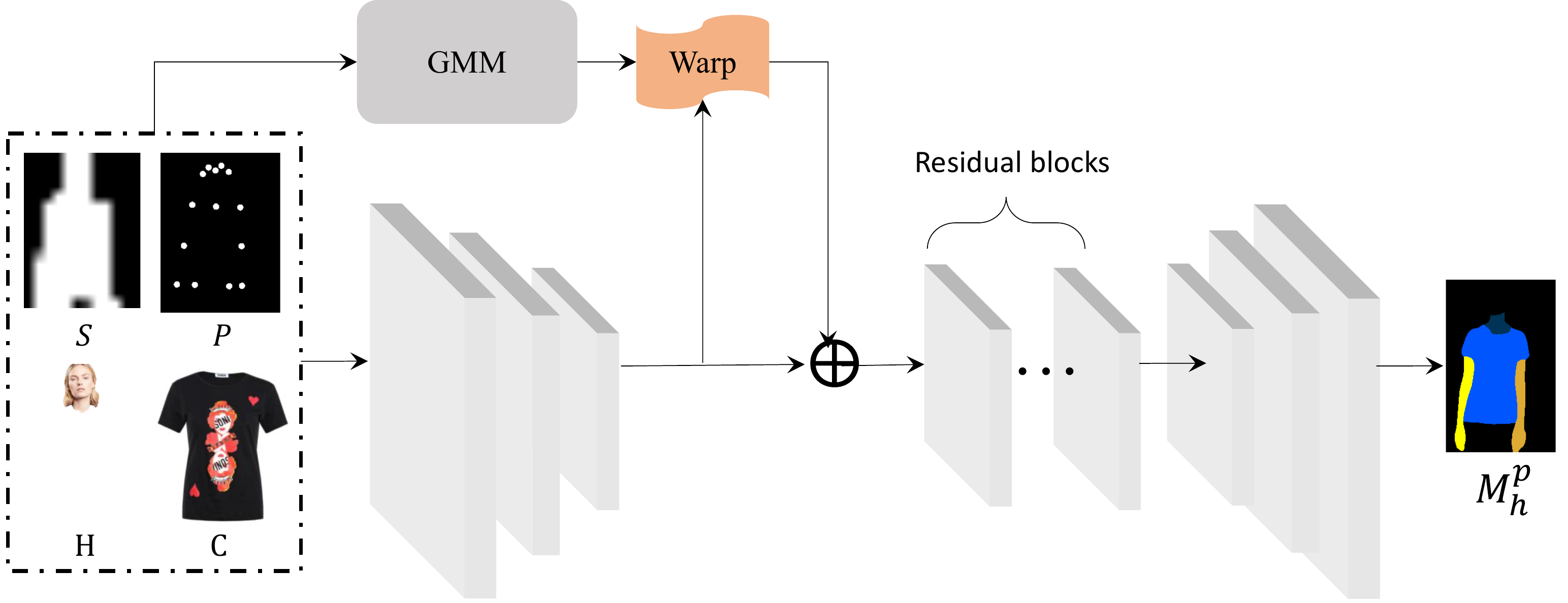}
  \vspace{-6mm}
  \caption{Architecture of the partial parsing prediction module. This module takes the body shape $S$, the person pose $P$, the head image $H$, and the flat clothing $C$ as inputs to predict the partial human parsing map $M_{h}^{p}$.}
  \vspace{-4mm}
  \label{fig:parsing_framework}
\end{figure}

\subsection{Partial Parsing Prediction Module}\label{parsing-predict}
We introduce the preparatory Partial Parsing Prediction (PPP) module to estimate the semantic label of the region to be changed during the try-on process.
The inputs to the PPP module consist of the blurred body shape $S$, the 18-channels pose heat map $P$ (obtained by applying~\cite{openpose} on the person image), the head image $H$ (obtained from the person parsing~\cite{Gong_2019_CVPR}) and the flat clothing image $C$.

As shown in Fig.~\ref{fig:parsing_framework}, our PPP module consists of an encoder-decoder architecture, where six residual blocks are used in the bottleneck and are defined to be part of the decoder. 
However, directly predicting the semantic label accurately is challenging as there is no guidance about how to warp the flat clothing into the target shape. 
To provide the model with more information about the target shape, the PPP module utilizes the Geometry Matching Module (GMM) from CP-VTON~\cite{bochao2018cpvton} to estimate the parameters of the TPS~\cite{bookstein1989TPS} transformation and warp the clothing features produced by the encoder. The pose-specific warped features are then concatenated with the clothing features and passed to the decoder to estimate the partial human parsing in the target pose.
It is worth noting that, different from the clothing warping module focusing on the texture details of the warped result, only the clothing shape matters in the PPP module.

During training, we use the pre-trained GMM from CP-VTON and fix its parameters. Further, the discriminator from pix2pixHD~\cite{wang2018pix2pixHD} is adopted to distinguish the synthesized human parsing results from the real one.
We utilize the adversarial loss in~\cite{mao2017lsgan} for the generator and the discriminator, which can be formulated as:
\begin{equation}
\mathcal{L}_{\text{adv}}^G=(D(c,G(c))-1)^2,
\label{eq:parsing_adv_g}
\end{equation}
\begin{equation}
\mathcal{L}_{\text{adv}}^D=(D(c,x)-1)^2+(D(c,G(c)))^2,
\label{eq:parsing_adv_d}
\end{equation}
where $G$ is the generator, $D$ is the discriminator, $x$ is the real partial parsing obtained by applying~\cite{Gong_2019_CVPR} on the real person image, and $c$ represents the inputs of the network.
To stabilize the adversarial training and accelerate model convergence, we introduce a feature matching loss $\mathcal{L}_{\text{fm}}$~\cite{wang2018pix2pixHD} between the real parsing features and the synthesized parsing features, which can be formulated as:
\begin{equation}
    \mathcal{L}_{\text{fm}} = \sum_{i=1}^T\frac{1}{N_i}\|D^{(i)}(c,x)-D^{(i)}(c,G(c))\|_1,
\end{equation}
where $T$ denotes the total number of layers in the discriminator and $N_i$ denotes the number of elements in the $i$-th layer. In addition, we also adopt the pixel-wise cross-entropy loss $\mathcal{L}_{\text{pixel}}$~\cite{gong2017lip} for the estimated parsing map. The total loss function can be formulated as (the trade-off hyperparameter $\lambda_{\text{adv}}$ is set to 0.1):
\begin{equation}
    \mathcal{L}_{\text{parsing}}^G =\mathcal{L}_{\text{pixel}} + \mathcal{L}_{\text{fm}} + \lambda_{\text{adv}}\mathcal{L}_{\text{adv}}^G,
\end{equation}
\begin{equation}
    \mathcal{L}_{\text{parsing}}^D = \mathcal{L}_{\text{adv}}^D.
\end{equation}

\begin{figure*}[t]
  \centering
  \includegraphics[width=1.0\hsize]{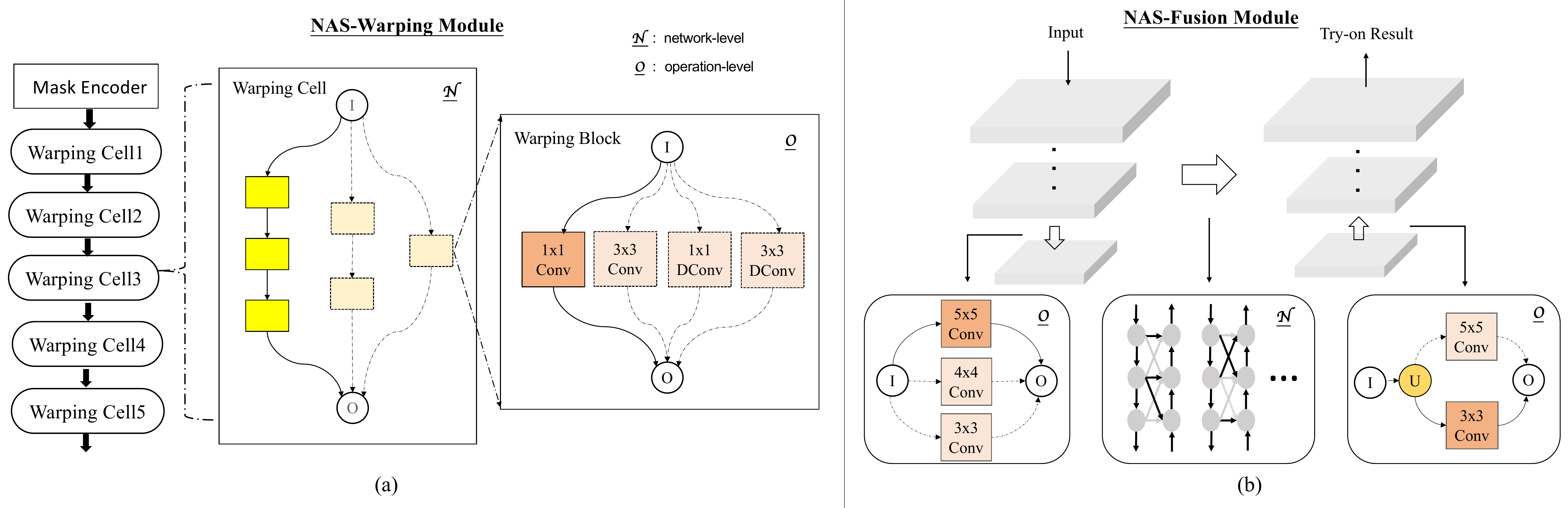}
  \vspace{-8mm}
  \caption{The search space for the NAS-Warping and the NAS-Fusion modules. Both of them have a network-level search space and an operation-level search space, labeled by \underline{\chancery{N}} and \underline{\chancery{O}}. For NAS-Warping, the network-level search space contains three warping branches within one warping cell, each of which has 1/2/3 warping blocks, and the operation-level search space designed for each warping block contains four operations: (1) 1$\times$1 convolution; (2) 1$\times$1 depthwise-separable convolution; (3) 3$\times$3 convolution; (4) 3$\times$3 depthwise-separable convolution. For the NAS-Fusion module, different types of skip-connections form the network-level search space, and two candidate operation sets for the upsampling and the downsampling processes respectively contain: (1) 3$\times$3 convolution; (2) 4$\times$4 convolution; (3) 5$\times$5 convolution; and (1) bilinear interpolation + 3$\times$3 convolution; (2) bilinear interpolation + 5$\times$5 convolution.}
  \label{fig:nas}
  \vspace{-4mm}
\end{figure*}

\subsection{NAS-Warping Module}\label{nas-warp}
To obtain the precise warping results under various deformation required for different clothing categories,
our NAS-Warping module is based on the flow warping method and searches category-specific warping networks.
As shown in Fig.~\ref{fig:nas}(a), we construct the supernet of the NAS-Warping module by one mask encoder and five multi-scale warping cells. 
The mask encoder aims to extract the source and target clothing features ($R_s$, $R_t$) from the corresponding clothing masks ($M_c$, $\widetilde{M_c}$), where $M_c$ is derived from the flat clothing $C$ and $\widetilde{M_c}$ comes from the ground truth parsing of the reference person during training or the result of the PPP Module during testing. 
The five warping cells are used to predict the clothing flow and warp the source features $R_s$ in a progressive manner. Note the architecture of the mask encoder in Fig.~\ref{fig:nas}(a) is fixed and only the five warping cells are searchable. 
The network-level search space \underline{\chancery{N}} (see Fig.~\ref{fig:nas}.(a)) is designed to take the variation in warping complexity that is required for different clothing types into account (see Fig.~\ref{fig:defomation_comparison}). Therefore, it explicitly controls the warping complexity by controlling the number of warping blocks (1/2/3) in each warping cell, making the searched warping network clothing category-specific.

Specifically, in each warping block, the source and target feature maps ($R_s$, $R_t$) are concatenated and passed to a convolution layer to estimate a flow map $F$. Then, the source feature $R_s$ is deformed by $F$ to the warped features $\widetilde{R}_s$, which will then be used in the next warping block to estimate another clothing flow. The operations in each warping block can be formulated as:
\begin{equation}
F = \mathcal{C}(R_s \oplus R_t),
\label{eq:warping_1}
\end{equation}
\begin{equation}
\widetilde{R}_s = \mathcal{W}(R_s,F),
\label{eq:warping_2}
\end{equation}
\begin{equation}
\widetilde{F} = \mathcal{W}(\widetilde{F},F),
\label{eq:warping_3}
\end{equation}
where $\oplus$ denotes the channel-wise concatenation, $\mathcal{C}(\cdot)$ represents the convolution operation, and $\mathcal{W}(\cdot,F)$ is the warping according to the estimated flow $F$. As shown in Eq.~\ref{eq:warping_3}, the estimated flow $F$ in the current warping block is also used to refine the flow $\widetilde{F}$ estimated in the previous warping block, and the final optimized flow will be directly leveraged for warping the flat clothing $C$ to the warped clothing $\widetilde{C}$, as shown in Fig.~\ref{fig:framework}. 

As for the operation-level search space\underline{\chancery{O}}, we provide a predefined convolution set (see Fig.~\ref{fig:nas}.(a) for details) for each warping block to search the optimal size of filters for a particular warping layer.

During training the supernet, we adopt the $\mathcal{L}_1$ loss $\mathcal{L}_{\text{mask}}$ between the warped clothing mask $\widetilde{M_c'}$ and the target clothing mask $\widetilde{M_c}$ to restrict the shape of the flow,
\begin{equation}
    \mathcal{L}_{\text{mask}} = \|\widetilde{M_c'}-\widetilde{M_c}\|_1.
\end{equation}
We also use the perceptual loss~\cite{johnson2016perceptual} $\mathcal{L}_{\text{perc}}$ between the warped clothing $\widetilde{C}$ and its ground truth $C_t$,  
\begin{equation}
\mathcal{L}_{\text{perc}}(\widetilde{C},C_t) = \sum_{k=0}^5 \lambda_{k}\|\phi_{k}(\widetilde{C})-\phi_{k}(C_t)\|_1,
\label{eq:warping_loss}
\end{equation}
where $\phi_k(I)$ denotes the $k$-th feature map in a VGG-19 network pre-trained on the ImageNet~\cite{ILSVRC15} dataset and $C_t$ can be obtained by applying the human parsing method~\cite{Gong_2019_CVPR} on the reference person image.
Finally, we introduce the TV loss~\cite{zach2007tv} $\mathcal{L}_{\text{TV}}$ of the estimated flow to avoid excessive distortion in the warped clothing, 
\begin{equation}
    \mathcal{L}_{\text{TV}} = \|\nabla F_x \|_1 + \|\nabla F_y \|_1,
\end{equation}
where $\nabla F_x$ and $\nabla F_y$ represent the difference of the flow map between adjacent positions in the x-axis and y-axis, respectively.
The total loss function can be formulated as:
\begin{equation}
     \mathcal{L}_{\text{NAS\\-Warping}} = \mathcal{L}_{\text{mask}} + \lambda_{\text{perc}}\mathcal{L}_{\text{perc}} + \lambda_{\text{TV}}\mathcal{L}_{\text{TV}},
\end{equation}
where $\lambda_{\text{perc}}$ and $\lambda_{\text{TV}}$ are the trade-off hyperparameters and are set to 0.1 and 0.3, respectively. 

\begin{table*}
    \centering
    \def\arraystretch{0.9}
    \tabcolsep 20pt
    \begin{tabular}{c|cc|cc|c}
        \toprule
        \multirow{2}*{Method} & \multicolumn{2}{c|}{warped clothes} & \multicolumn{2}{c|}{try-on results} & \multirow{2}*{user study score} \\
        \cmidrule{2-5}
        & SSIM $\uparrow$ & FID $\downarrow$ & SSIM $\uparrow$ & FID $\downarrow$ \\
        \midrule
        VITON~\cite{xintong2018viton} & 0.7171 &  66.56 & 0.7056 & 44.44 & 5.05\% \\
        CP-VTON~\cite{bochao2018cpvton} &0.7469 & 59.84 & 0.7354 & 30.99 & 10.21\% \\
        ACGPN~\cite{han2020acgpn} & 0.8099 & 35.91 &  0.8357 & 23.49 & 21.67\% \\
        \midrule
        WAS-VTON (Our) & \textbf{0.8577} & \textbf{28.21} & \textbf{0.8430} & \textbf{13.83} & \textbf{63.07\%}\\
        \bottomrule
    \end{tabular}
    \caption{The SSIM~\cite{2004SSIM} score and FID~\cite{heusel2017fid} score among different methods on the extended virtual try-on dataset~\cite{xintong2018viton} as well as the user evaluation results.}
    \label{tab:quantitative result}
    \vspace{-8mm}
\end{table*}

\subsection{NAS-Fusion Module}\label{nas-label}
To seamlessly fuse the warped clothing and the person image,
our NAS-Fusion module searches a network with particular skip-connections between the same or different scales of features to deliver the most instructive information from the encoder to the decoder. 
Before discussing the search space, we first introduce its overall framework. 
As shown in Fig.~\ref{fig:framework}, the NAS-Fusion module takes the concatenation of the preserved person image $I'$ extracted from the real person image $I$ by parsing (head, hair and lower-body (resp., upper-body) in case of upper-body (resp., lower-body) try-on), the warped clothing $\widetilde{C}$, the human parsing map $M_h$ (obtained by combining the $M_h^p$ and the full parsing of the preserved person image), and the 18-channel pose heat map $P$ as inputs to synthesize the final try-on result $\widetilde{I}$. 
More specifically, the fusion module is an encoder-decoder network
aiming to predict the coarse try-on result $I_c$ and the fusion mask $M_f$.
The predicted fusion mask $M_f$ is used to fuse the warped clothing $\widetilde{C}$ and the coarse result $I_c$ to obtain the final try-on result $\widetilde{I}$. The fusion process can be formulated as:
\begin{equation}
    \widetilde{I} = I_c \odot (1-M_f) + \widetilde{C} \odot M_f.
\end{equation}

As shown in Fig.~\ref{fig:nas}(b), the supernet comprises candidate architectures with various skip-connections among features between encoder and decoder, and different convolution operations for each network layer. 
Each decoding layer can be skip-connected to the same, the previous, or the next hierarchy level in the encoder. The NAS-Fusion module aims to search the optimal skip-connection for each network layer in the decoder and optimal convolution operations of the downsampling and upsampling process. 

During the training of the supernet, we utilize the $\mathcal{L}_1$ loss and the perceptual loss~\cite{johnson2016perceptual} $\mathcal{L}_{\text{perc}}$, which can be formulated as:
\begin{equation}
\begin{split}
    \mathcal{L}_{\text{NAS-Fusion}} = &\mathcal{L}_1(I,I_c)+\mathcal{L}_{\text{perc}}(I,I_c)+\mathcal{L}_1(I,\widetilde{I}) \\
    &+\mathcal{L}_{\text{perc}}(I,\widetilde{I})+\mathcal{L}_1(\widetilde{M_c}, M_f).
\end{split}
\end{equation}

\subsection{Network Optimization and Architecture Search}
To facilitate a stable and time-efficient NAS, our WAS-VTON inherits the one-shot framework in~\cite{zichao2019singlepath} to automatically search the optimal architecture for the NAS-Warping module and the NAS-Fusion module. For each module, the supernet training and the architecture search are conducted sequentially.
Specifically, in each training iteration of the NAS-Warping supernet, for each warping cell, we randomly sample one warping branch that has a particular number (1/2/3) of warping blocks, and then sample one convolution operation for each warping block, as shown in Fig.~\ref{fig:nas}(a). 
Similarly, in each training iteration of the NAS-Fusion supernet, for each decoder layer, we randomly sample one of the three types of skip-connection to get features from the encoder. Additionally, we also randomly sample an operation from the corresponding candidate set for each network layer in the encoder and the decoder.
(see Fig.~\ref{fig:nas}(b) for more details). 
Note that during training of the supernet for the NAS-Warping module or the NAS-Fusion module, clothes from all five categories 
(vest/sling, skirt, pants, short- and long-sleeve shirt in our dataset) are fed into the network.

After training the supernet, an evolutionary algorithm~\cite{ea} is applied to search the optimal architecture and the Structure Similarly Index Measure (SSIM)~\cite{2004SSIM} score is used to supervise the searching process, i.e., SSIM between the warped clothing and the clothing-on-person for the NAS-Warping module, and SSIM between the synthesized person and the real person image for the NAS-Fusion module.
For the NAS-Warping search of a particular clothing category, the evolutionary algorithm is conduct independently by using the data from the corresponding category.
For the NAS-Fusion search, instead, all clothing types are used in order to find the most useful category-agnostic fusion architecture to fuse the warped clothing and the unchanged person part.


\section{Experiments}
\subsection{Dataset and Implementation Details}
\noindent\textbf{Dataset.}~\label{sec:dataset} 
We conduct comprehensive experiments on the VITON dataset~\cite{xintong2018viton} which contains 16,235 image pairs of front-view person and upper clothing. There are three types of clothing in the VITON dataset, namely, short sleeve, long sleeve, and vest (sling). To increase the clothing diversity, we follow the image crawling protocol in~\cite{xintong2018viton} and expand the original dataset with pants and skirts. The expanded dataset is split into a training and a testing set, with 18,312 and 2,487 image pairs, respectively. 
During network searching, we further split the testing set to obtain a validation set containing 950 image pairs, which is not used for the final evaluation. 

\noindent\textbf{Implementation Details.} 
We train the PPP module, the NAS-Warping supernet, and the NAS-Fusion supernet for 9, 200, and 270 epochs, and the batch sizes are set to 20, 10, and 10. During training, each module uses 2, 1, and 1 NVIDIA GTX 1080Ti GPUs, respectively. The Adam optimizer~\cite{Kingma2014adam}  with $(\beta_1 = 0.5, \beta_2 = 0.999)$ is used for all modules and the initial learning rates are set to 0.002, 0.0002, 0.0001, respectively. When using the evolutionary algorithm for network searching, we set the max iterations to 25, the population size to 40, the crossover number to 15, and the mutation number to 15.
With the experiment setting above, training the PPP module takes about 1.5 hours. Training the NAS-Warping supernet takes about 3 days while the time for searching the clothing category-specific warping networks varies from 3 hours to 11 hours as the amount of data varies with clothes category. Training the NAS-Fusion supernet takes about 2.5 days while the searching procedure takes about 16 hours. Once we obtain the searched warping networks and the searched fusion network, the inference time is on-par with the existing virtual try-on approaches.

\begin{figure*}[t]
  \centering
  \includegraphics[width=1.0\hsize]{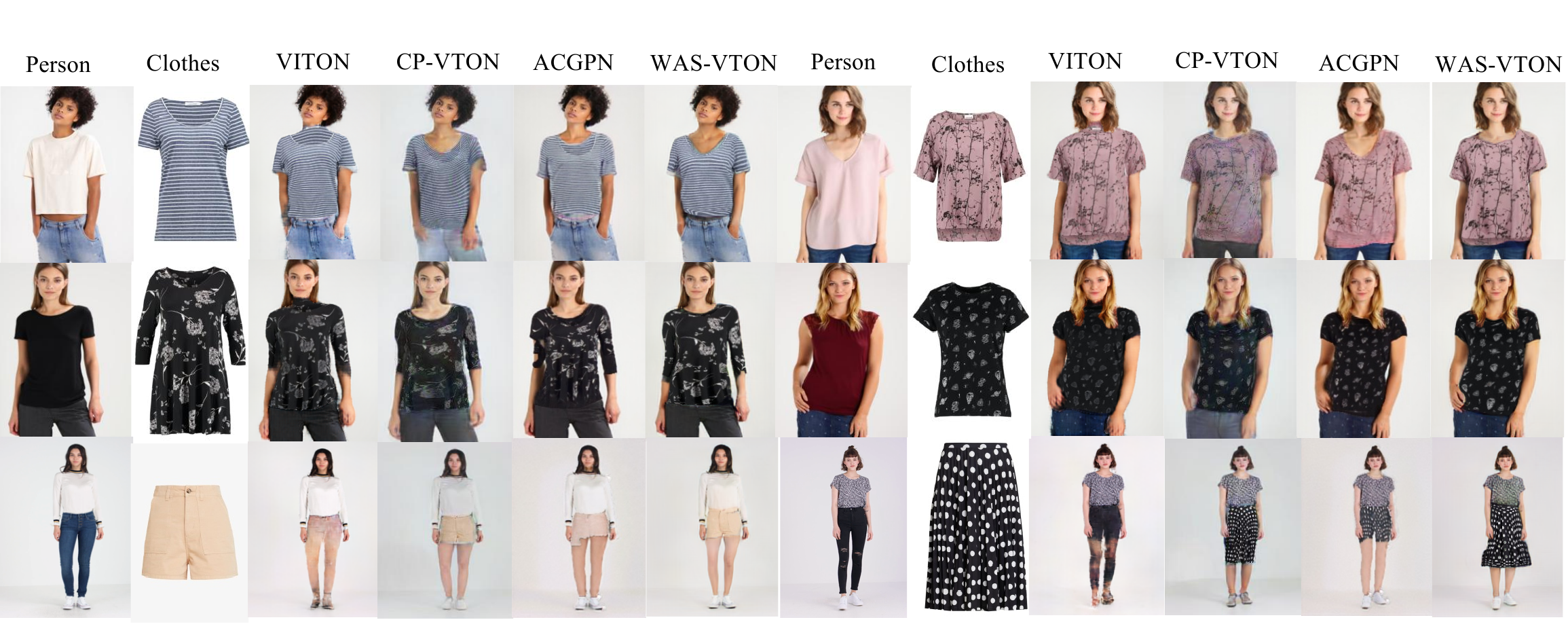}
  \vspace{-8mm}
  \caption{Qualitative comparison results among the proposed WAS-VTON and other approaches. (Zoom in for more details.) }
  \vspace{-4mm}
  \label{fig:fig4}
\end{figure*}

\begin{table}
    \centering
    \tabcolsep 2pt
    \begin{tabular}{c|ccc|c}
         \toprule
         Score & VITON~\cite{xintong2018viton} & CP-VTON~\cite{bochao2018cpvton} & ACGPN~\cite{han2020acgpn} & WAS-VTON \\
        \midrule
        SSIM $\uparrow$ & 0.7056 &  0.7171 & 0.8099 & \textbf{0.8306}  \\
        FID $\downarrow$ & 44.47 & 38.02 & 29.00 &  \textbf{13.71}  \\
        \bottomrule
    \end{tabular}
     \caption{The SSIM~\cite{2004SSIM} score and FID~\cite{heusel2017fid} score among different methods on the original virtual try-on dataset~\cite{xintong2018viton}. 
      } 
    \label{tab:original_comp}
    \vspace{-8mm}
\end{table}

\subsection{Quantitative Comparison with SOTAs}
We quantitatively compare our WAS-VTON with three existing virtual try-on methods, VITON~\cite{xintong2018viton}, CP-VTON~\cite{bochao2018cpvton}, and ACGPN~\cite{han2020acgpn}. The SSIM~\cite{2004SSIM} and the Fr$\mathbf{\acute{e}}$chet Inception Distance (FID)~\cite{heusel2017fid} are used to evaluate the warping and the final try-on results.
As shown in Table~\ref{tab:quantitative result}, our proposed WAS-VTON outperforms the other methods, evidencing that our WAS-VTON can warp clothes more accurately and synthesize more realistic try-on results. 
We also conduct a user evaluation study
to assess the quality of the try-on results. 63.07\% of the users consider the results from WAS-VTON more realistic, further demonstrating the superiority of the proposed WAS-VTON. 
To fairly compare with the existing methods, we further test our WAS-VTON on the original virtual try-on dataset~\cite{xintong2018viton} without expanded data. As shown in Table~\ref{tab:original_comp}, our WAS-VTON is still the best-performing approach on the original dataset.

In addition, we also perform extensive ablation experiments to verify the effectiveness of the NAS-Warping and the NAS-Fusion module. We compare the widely used DARTS~\cite{liu2018dasrts} searching framework with the single-path method adapted for our WAS-VTON in Table~\ref{tab:compare_darts}, showing the superiority of the singe-path paradigm.
Further, in Table~\ref{tab:ablation}, we first design three ablation baselines for the NAS-Warping module, each of which contains a fixed number (1/2/3) of warping blocks in each warping cell, with the operation in each warping block being a 3$\times$3 convolution. Compared with those three baselines, our searched NAS-Warping network can obtain the highest SSIM and a competitive FID score for the clothing warping result.
Given the NAS-Warping network, we also conduct an ablation on the NAS-Fusion module. The "w/o dynamic skips" in Table~\ref{tab:ablation} means that the U-Net-like architecture is directly set as the fusion network, in which the operation of every downsampling and upsampling layer is fixed to a 3$\times$3 convolution. 
By adding the dynamic skip connections, and thereby utilizing the searched NAS-Fusion network, the best SSIM (0.8430) and FID (13.83) scores can be obtained for the final try-on results.
\begin{table}
    \centering
    \tabcolsep 8pt
    \begin{tabular}{c|cc|cc}
        \toprule
        \multirow{2}*{Method} & \multicolumn{2}{c|}{warped clothes} & \multicolumn{2}{c}{try-on results}\\
        \cmidrule{2-5}
        & SSIM $\uparrow$ & FID $\downarrow$ & SSIM $\uparrow$ & FID $\downarrow$ \\
        \midrule
        DARTS~\cite{liu2018dasrts} & 0.8469 &  33.40 & 0.8025 & 26.93 \\
        Our & \textbf{0.8577} & \textbf{28.21} & \textbf{0.8430} & \textbf{13.83} \\
        \bottomrule
    \end{tabular}
    \caption{Quantitative comparison between the DARTS~\cite{liu2018dasrts} and the single-path (Ours) searching results.}
    \label{tab:compare_darts}
    \vspace{-10mm}
\end{table}

\begin{table}[]
\centering
\begin{tabular}{c|c|cc}
\toprule
\multirow{5}{*}{NAS-Warp} & \# warp. blocks & SSIM $\uparrow$  & FID $\downarrow$ \\ \cmidrule{2-4}
                     & 1              & 0.8556       & 28.33   \\
                     & 2              & 0.8525       & 29.25 \\
                     & 3              & 0.8565       & \textbf{27.79}   \\
                     & Ours              & \textbf{0.8577}    & 28.21   \\
\midrule
\multirow{3}{*}{NAS-Fusion} & dynamic skips & SSIM $\uparrow$  & FID $\downarrow$ \\ \cmidrule{2-4}
                     & \xmark         & 0.8385  & 16.79  \\    
                     & \cmark         & \textbf{0.8430}  & \textbf{13.83}  \\
\bottomrule
\end{tabular}
\caption{The ablation study on the NAS-Warping and the NAS-Fusion modules.}
 \vspace{-7mm}
\label{tab:ablation}
\end{table}

\begin{figure}
  \centering
  \includegraphics[width=1.0\hsize]{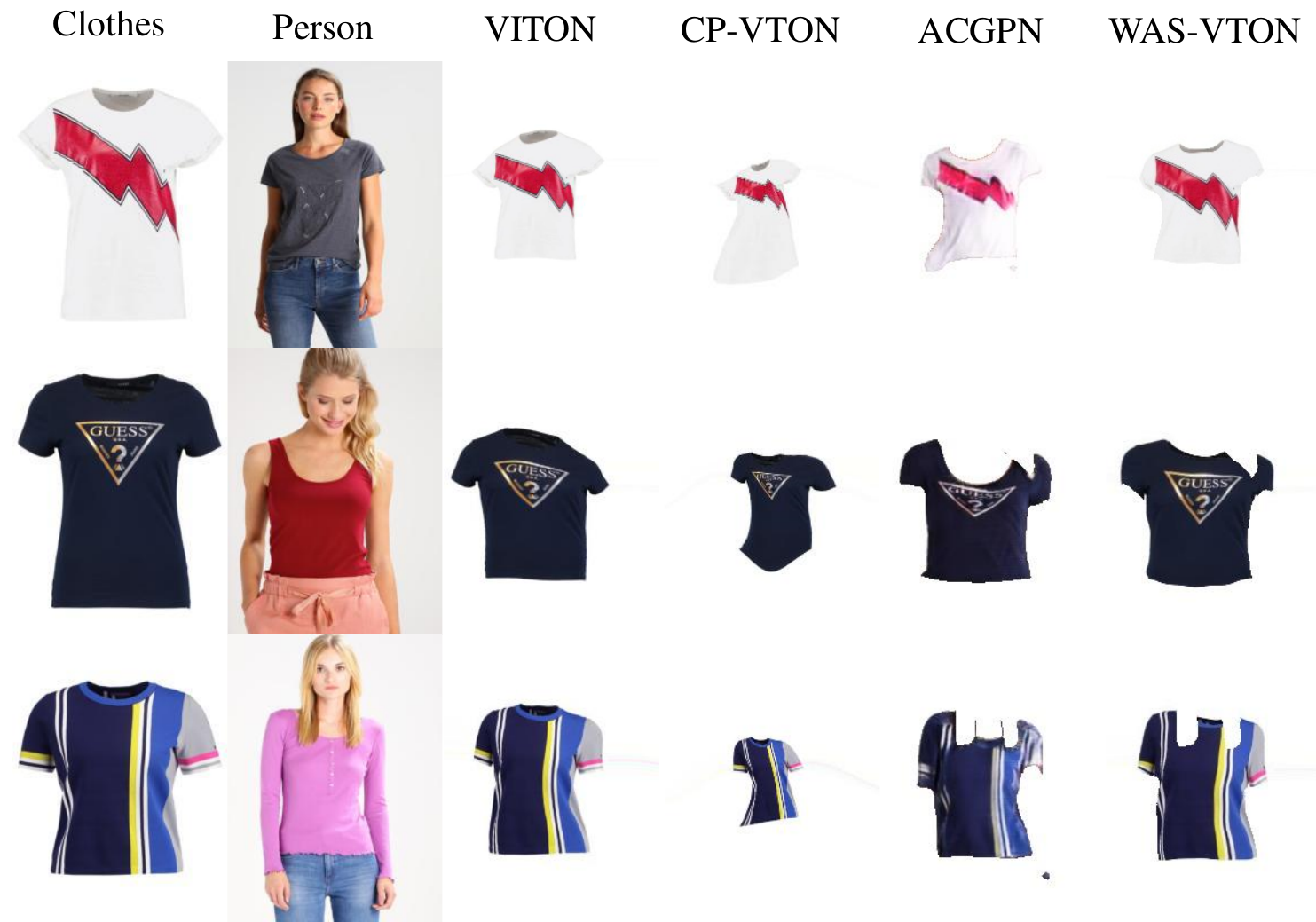}
  \vspace{-8mm}
  \caption{Visual comparison of warped clothes among our WAS-VTON and other approaches.}
  \vspace{-4mm}
  \label{fig:fig6}
\end{figure}

\begin{figure*}[t]
  \centering
  \includegraphics[width=1.0\hsize]{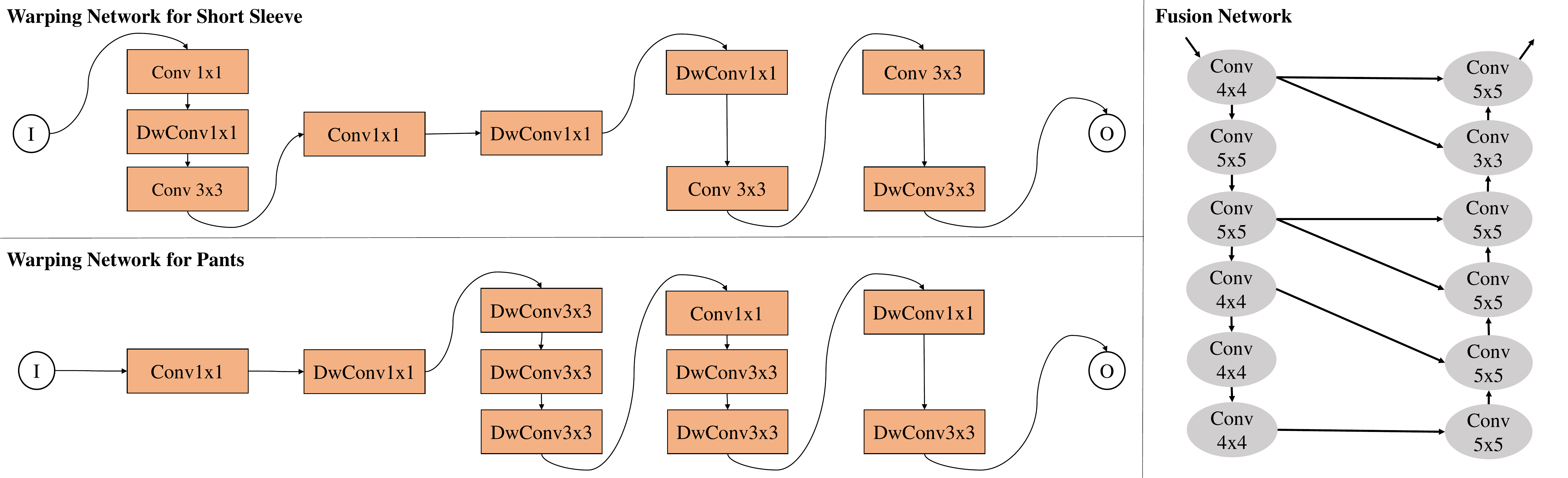}
  \vspace{-7mm}
  \caption{Searched architectures for the NAS-Warping and the NAS-Fusion modules. 
  \vspace{-2mm}
  }
  \label{fig:fig5}
\end{figure*}

\begin{table*}[t]
    \centering
    \tabcolsep 20pt
    \begin{tabular}{c|ccccc|c}
    \toprule
         \multirow{2}*{Clothes Category} & \multicolumn{5}{c|}{Resolution} & \multirow{2}*{FLOPs} \\
         \cmidrule(r){2-6}
         & 12$\times$16 & 12$\times$16 & 24$\times$32 & 48$\times$64 & 96$\times$128 \\
         \midrule
         Short Sleeve & (0,2,1) & (0) & (2) & (2,1) & (1,3) & 876.5M \\
         Long Sleeve & (0,1,1) & (0) & (0) & (3,2,3) & (3,3,3) &  1006.6M \\
         Sling \& Vest & (0,0,1) & (3,1,1) & (0) & (1,1) & (1,3) & 986.6M \\
         Pants & (0) & (2) & (3,3,3) & (0,3,3) & (1,3) & 934.7M \\
         Skirt & (0,0,1) & (2,1) & (0) & (3,1,3) & (3,3,3) & 
1102.1M \\
        \bottomrule
    \end{tabular}
    \caption{The searched architectures for category-specific warping modules. The label 0, 1, 2, 3 represent 1$\times$1 convolution, 3$\times$3 convolution, 1$\times$1 depthwise-separable convolution, and 3$\times$3 depthwise-separable convolution, respectively.}
    \vspace{-8mm}
\label{tab:naswarping}
\end{table*}

\subsection{Qualitative Comparison with SOTAs}
Fig.~\ref{fig:fig4} presents a visual comparison of WAS-VTON with VITON\cite{xintong2018viton}, CP-VTON\cite{bochao2018cpvton}, and ACGPN\cite{han2020acgpn}. 
As shown in Fig.~\ref{fig:fig4}, VITON usually results in unrealistic warping results including visual artifacts, distorted texture, and unreasonable clothes silhouette. Though CP-VTON can alleviate those issues to some extent, the texture details are usually blurred. 
Further, both VITON and CP-VTON tend to fail when the clothes are occluded by some body part. 
ACGPN does better in retaining cloth texture as well as the unchanged person appearance, and with the guidance of the human parsing, it can synthesize reasonable try-on results in the occlusion scenarios. However, the edge region of the clothes in ACGPN is still blurred and there are obvious artifacts at the collar region. Moreover, all the above three methods perform poorly for the lower-body try-on.
WAS-VTON, instead, is able to preserve the clothing texture and other parts of the body well, resulting in high-quality full-body try-on results, especially in the collar and clothing silhouette regions. Thanks to the PPP module and the clothing category-specific warping network, WAS-VTON also performs well in the presence of occlusion and obtains more compelling warping results (see Fig.~\ref{fig:teaser} and Fig.~\ref{fig:fig6}).

\subsection{Searched Architecture Analysis}
Table~\ref{tab:naswarping} shows the optimal warping networks for different clothing categories.
We can observe some patterns in the learned warping networks. First, the network architectures for various clothing categories are different, verifying our hypothesis that clothes from different categories require different warping networks. 
Second, all networks tend to utilize convolutions with small kernel size (1$\times$1) for low-resolution features while preferring convolutions with larger kernel sizes (3$\times$3) for high-resolution features.
Evidenced by the quantitative comparison in Table~\ref{tab:quantitative result}, it can be argued that using fixed convolutions for all features is therefore not the best choice for the warping network. Third, the learned network structure prefers depthwise-separable convolutions over normal convolutions for features with high-resolution. Fig.~\ref{fig:fig5} shows two searched warping networks and the searched fusion network. Different from the normal UNet used for image translation, the fusion network for the try-on task leverages skip-connections between the high-resolution encoding features and the low-resolution decoding features. The NAS-Fusion module automatically searches the skip-connections for each decoding layer to make full use of the geometric information like clothing silhouettes (resp., the semantic information) embedded in high-resolution (resp., low-resolution) encoding feature maps, which can better facilitate synthesis of the final try-on result.

\section{Conclusion}
Taking a step forward from the existing virtual try-on methods, we investigate the network architecture search (NAS) for the virtual try-on task and propose a dynamic Warping Architecture Search for Virtual Try-ON (WAS-VTON), including the PPP module, the NAS-Warping module, and the NAS-Fusion module. The PPP module estimates the semantic label of the clothing region to be changed. The NAS-Warping module searches category-specific warping networks for different kinds of clothes while the NAS-Fusion module searches a fusion network with a particular skip-connection configuration to fuse the warped clothes and person image more seamlessly. Besides, we expand the existing virtual try-on dataset to increase clothing category diversity. Extensive experiments illustrate the effectiveness of our WAS-VTON and show its superiority over existing state-of-the-art architecture-fixed try-on methods.

\section{Acknowledgments}
This work was supported in part by National Key R\&D Program of China under Grant No. 2020AAA0109700, National Natural Science Foundation of China (NSFC) under Grant No.U19A2073 and No.61976233, Guangdong Province Basic and Applied Basic Research (Regional Joint Fund-Key) Grant No.2019B1515120039, Guangdong Outstanding Youth Fund (Grant No. 2021B1515020061), Shenzhen Fundamental Research Program (Project No. RCYX202007 14114642083, No. JCYJ20190807154211365), Zhejiang Lab’s Open Fund (No. 2020AA3AB14) and CSIG Young Fellow Support Fund.

\clearpage

\bibliographystyle{ACM-Reference-Format}
\bibliography{mfp1768}

\end{document}